%% file: eccv2020submission.tex
\documentclass[runningheads]{llncs}
\usepackage{graphicx}
\usepackage{comment}
\usepackage{amsmath,amssymb} 
\usepackage{color}

\usepackage{epsfig}
\usepackage{graphicx}
\usepackage{amsfonts}
\usepackage{mathrsfs}
\usepackage{booktabs}
\usepackage{xcolor}

\usepackage[bottom]{footmisc}

\newcommand{\etal}{\textit{et al}.}

\font\dsrom=dsrom10 scaled 1200
\newcommand{\indicator}[1]{\textrm{\dsrom{1}}_{#1}}

\begin{document}

\pagestyle{headings}
\mainmatter
\def\ECCVSubNumber{4246}

\title{
  Learning to Visually Navigate in Photorealistic Environments Without any Supervision
}

\titlerunning{
  Learning to Visually Navigate Without any Supervision
}

\author{
  Lina Mezghani\inst{1,2}
  \and
  Sainbayar Sukhbaatar\inst{1}
  \and
  Arthur Szlam\inst{1}
  \and \\
  Armand Joulin\inst{1}
  \and
  Piotr Bojanowski\inst{1}
}

\institute{
  Facebook AI Research 
  \and
  INRIA 
  \footnote{Univ. Grenoble Alpes, Inria, CNRS, Grenoble INP, LJK, 38000 Grenoble, France}
}

\authorrunning{Mezghani et al.}

\maketitle

\begin{abstract}
Learning to navigate in a realistic setting where an agent must rely solely on visual inputs is a challenging task, in part because 
the lack of position information makes it difficult to provide supervision during training.
In this paper, we introduce a novel approach for learning to navigate from image inputs without external supervision or reward.
Our approach consists of three stages: learning a good representation of first-person views, then learning to explore using memory, and finally learning to navigate by setting its own goals. 
The model is trained with intrinsic rewards only so that it can be applied to any environment with image observations.
We show the benefits of our approach by training an agent to navigate challenging photo-realistic environments from the Gibson dataset~\cite{xiazamirhe2018gibsonenv} with RGB inputs only.
\end{abstract}


\input{introduction}

\input{related}


\input{problem}


\input{model}


\input{experiments}


\section{Conclusion}
We have shown how to train an agent to perform goal-directed navigation in photorealistic environments without using any extrinsic rewards.
Our agent trains in a purely self-supervised manner,
only using RGB image observations.
The model is composed of three interconnected components:  one which learns visual representations, a second which explores the environment, and a third that teaches itself to navigate.
We have shown that our self-supervised navigation models manage to navigate to novel test goals.

In future work, we can consider multiple natural extensions to this model.
First, we want to work on making the learning of all the components of the model end-to-end.
Second, we want to study the generalization capabilities of our method by training it on a large set of scenes with shared parameters and testing it on previously unseen environments. 
Finally, we can improve the dependency on the SMB by including the graph structure in the attention mechanism for both exploration and navigation policies.

\section*{Acknoledgements}
We would like to thank Dhruv Batra, Oleksandr Maksymets, Danielle Rothermel and Herv\'e J\'egou for their invaluable help and constructive comments throughout this project.

\bibliographystyle{splncs04}
\bibliography{egbib}
\end{document}

%% file: introduction.tex
\section{Introduction}
Designing algorithms for learning to navigate is a classical problem in robotics.  
This problem is challenging, especially in settings where it is necessary to do without accurate depth or position information; or more generally, with as little supervision as possible.
Furthermore, if the goal location is specified as an image, the agent needs to learn a good visual representation and an efficient exploration strategy in addition to the navigation policy.

One important set of approaches, called Simultaneous Localization And Mapping (SLAM)~\cite{thrun2005probabilistic}, builds a map of an environment {\it while} keeping track of where the agent is in the partial map.
Although many SLAM methods use statistical methods to improve estimation, until recently they did not emphasize statistical {\it learning}.
Thus these methods are unable to generalize and make use of regularities in the environment (or between environments) beyond what has been built by hand into the algorithm.

There has been a recent interest in using techniques from deep learning in the context of SLAM, or more generally, in the context of navigation~\cite{avraham2019empnet,fang2019scene,gupta2017cognitive,henriques2018mapnet,khan2017memory,oh2016control,parisotto2017neural,wierstra2010recurrent,zhang2017neural}.
Deep learning-based methods typically require a large number of trials during training and have been rarely considered outside of simulators.
However, the growing number of photo realistic environments~\cite{Matterport3D,xiazamirhe2018gibsonenv}, efficient simulators~\cite{Dosovitskiy17,habitat19iccv} and dedicated methods to transfer from simulated to real environments~\cite{sadeghi2016cad2rl,tobin2017domain} have fueled the research in deep learning-based navigation methods.

In a separate line of study, there has been great progress in learning image representations through ``self-supervised'' approaches \cite{caron2018deep,doersch2015unsupervised,goyal2019scaling,zhang2016colorful}.
In these works, using prior knowledge about the basic regularities of images,  researchers find pretext tasks that, when solved, give good feature representations for other tasks of interest.
While self-supervised learning is interesting for understanding learning methods abstractly, it also promises to be important in applications, as it is often the case that a pretext task is easier to come by and more general than strong supervision.

\begin{figure}[t]
  \centering
  \includegraphics[width=0.8\linewidth]{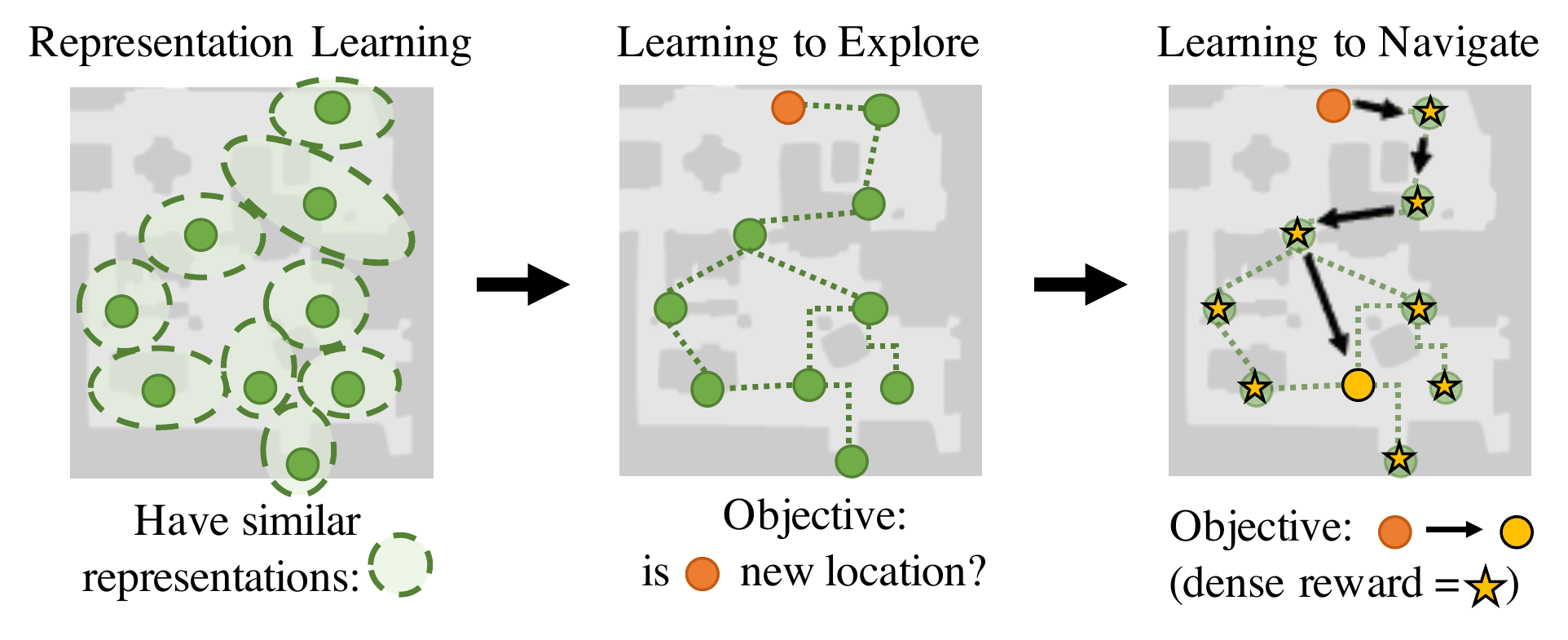}
  \vspace{-2mm}
  \caption{
\textbf{Three stages of training:}
the agent learns to distinguish locations from its visual inputs,
then it explores the environment and build a map of the environment,
finally it uses this map to learn how to navigate the environment.
Each step requires no external supervision or reward and the agent has only access to a visual RGB input, and has no information about its position.}
  \label{fig:pullfigure}
\end{figure}

In this work, we introduce an entirely unsupervised method for learning to navigate through simulators like Habitat~\cite{habitat19iccv} in photorealistic environments and large-scale three-dimensional point clouds such as the Gibson dataset~\cite{xiazamirhe2018gibsonenv}.
In particular, we assume that the agent only has access to image observations and that the target location is also given as an image.
The method is composed of three stages.
First, the agent learns a visual representation that can distinguish between nearby and far-away pairs of points in a similar way to~\cite{savinov2018episodic}.
The fundamental prior knowledge we use is that in most situations, an agent's representation of the world should not change very fast as it moves; but on the other hand, for most pairs of far-away points, the representations should be different.
Next, the agent learns to explore, adding states to a memory buffer when their feature representations are dissimilar to any in the buffer.
Finally, the agent trains itself to complete navigation tasks,  using the buffer to shape the reward for the navigation policy.
An important component of our model is that the agent uses a Scene Memory Buffer for both its policy and reward.
In particular, the agent takes actions via a Transformer~\cite{vaswani2017attention} applied to the memory buffer.
Because our approach can be applied in situations where the practitioner has no control over the environment - and in particular, with no ability to give supervision or move to arbitrary positions in the environment - the method is general.
We show that despite this generality, its final navigation policy outperforms other approaches.

\ \\

Our contributions in this paper are the following:
\begin{itemize}
  \item We propose a novel three-stage algorithm for learning to navigate using only RGB vision without any external supervision or reward in photorealistic environments that simulates actual houses.
  \item We introduce several improvements to the exploration policy~\cite{savinov2018episodic} such as conditioning on past memory and using discrete rewards. 
  \item We evaluate our model and show that it outperforms all baselines on scenes from the Gibson dataset.
\end{itemize}

%% file: related.tex
\section{Related Work}

Iteratively building a map of an environment to perform localization or navigation tasks has been extensively studied in robotics in the context of SLAM ~\cite{thrun2005probabilistic}.
Standard SLAM is composed of multiple hand-crafted modules to fit with the physical constraints of a robot~\cite{mur2015orb}.

Recently, several works have replaced components of SLAM with neural networks; for example, Chaplot~\etal~\cite{chaplot2018active} replace the localization module.
Gupta~\etal~\cite{gupta2017cognitive} propose a model composed of two successive modules, a mapper to build a latent world map, and a planner, that takes actions based on this map.
The mapper does not have dedicated external rewards but the planner performs tasks associated with external rewards and backpropagates the resulting gradients to the mapper.
This map has been further extended with image features~\cite{gupta2017unifying} or with a dynamic structure~\cite{avraham2019empnet,henriques2018mapnet}.
Other works replace SLAM entirely by deep models with no planning but explicit map-like or SLAM inspired memory structures~\cite{khan2017memory,oh2016control,parisotto2017neural,zhang2017neural}.
Closer to our work, Kumar~\etal~\cite{kumar2018visual} use human-made trajectories stored as sequences of feature representations of views, and Fang~\etal~\cite{fang2019scene} show the potential of the Transformer layer~\cite{vaswani2017attention} as a scene memory for navigating realistic environments.
As opposed to these works, our model is trained with intrinsic reward only.

Alternatively, several work train deep models to solve a navigation task without explicit world representations.
Mirowski~\etal~\cite{mirowski2016learning} learn a navigation policy with a recurrent network in synthetic mazes, and later, in real-world data from Google Maps~\cite{mirowski2018learning}.
Similar to our work, they use a surrogate loss on loop closure to help the training of the model, but they use sparse external reward to guide its training.
Similarly, Zhu~\etal~\cite{zhu2017target} show the benefit of deep models on a localization task framed as finding an observation taken from the goal.
Later, Yang~\etal~~\cite{yang2018visual} extend this to navigation to an object described only by its name.

Many works train agents to explore the world with an intrinsic reward~\cite{chentanez2005intrinsically,pathak2017curiosity,schmidhuber1991curious}.
For example, the curiosity-driven reward of Pathak~\etal~\cite{pathak2017curiosity} encourages agents to move to states that are hard to predict.
Of particular interest, Chen~\etal~\cite{chen2019learning} propose a coverage reward that encourages the agent to explore every part of its latent map.
This reward is quite general and benefits both exploration and navigation, but it does not directly optimize for navigation like ours.

Finally, our approach is most related to a recent line of research that uses multiple stages of learning to build a set or graph of scene observations \cite{eysenbach2019search,savinov2018semi,savinov2018episodic,zhang2018composable}.
Savinov \etal~\cite{savinov2018semi} internalize a landmark memory obtained from human trajectories.
They store representations of the locations visited in the trajectories and build a navigation graph based on their similarities.
Our work follows their self-supervised training of a reachability network $R$ to distinguish between nearby observations, but we extend the self-supervision to both exploration and navigation.
Savinov~\etal~\cite{savinov2018episodic} also use a curiosity-driven intrinsic reward based on a memory buffer.
Our exploration phase follows an intrinsic reward inspired by their work, but we also use the memory buffer in our Transformer-based policy.
Finally, Eysenbach~\etal~\cite{eysenbach2019search} propose a method to learn an agent to explore and navigate an environment with intrinsic rewards.
Their training follows the same sequence of steps as ours, with the exception that they clean the graph by testing existing edges and adding new ones and then learning to navigate on the graph, and not the environment.
Instead, our agent trains itself to navigate the environment directly by shaping dense rewards from the memory buffer.
It means that our agent can potentially learn more efficient navigation strategies not constrained to paths on the memory-graph.

%% file: problem.tex
\section{Problem formulation}

In this paper, we simulate a realistic setting where an agent must learn to navigate in a 3D environment.
We formulate this problem using the following assumptions:
\begin{itemize}
\item[--]\emph{No extrinsic reward.} We do not have control over the environment and thus cannot add extrinsic reward to guide the training of the agent.
\item[--]\emph{No human guidance.} The environment is new and has never been explored. We do not have access to human trajectories or other forms of external information. 
\item[--]\emph{3d scan environments.} We focus on photo-realistic environments such as the ones in the Habitat platform.
\end{itemize}
We are interested in the capability of the agent to explore and navigate an environment and we report the following metrics to measure its success:
\begin{itemize}
\item[--] \emph{Coverage.} We measure the coverage of an environment by an agent by discretizing the map into $C$ cells of the same size and counting the ratio $p_t$ of visited cells $c_t$ by the agent after $t$ steps.
\item[--] \emph{Image driven navigation.} We measure the capability of an agent to navigate the environment to an image target. That is: we give the agent an image observation from the location and we measure the number of steps it takes to reach the destination so that the agent's observation matches the image target, starting from the entry point of the map.
\end{itemize}
Finally, as a secondary goal, we are also interested in the robustness of an agent to limited sensor data. To that end, we focus on RGB inputs in this paper.
We do not use depth, gps coordinate or relative position as inputs.

%% file: model.tex
\section{Approach}
In this section, we describe our algorithm and its three stage training: first the agent learns a visual representation of the environment from random trajectories, then it learns to explore the environment to build a latent map, and finally it trains itself to navigate using the map.
Each step has a module trained without external supervision.

\subsection{Stage 1: Visual representation of the environment}
\label{sec:rnet}

\begin{figure}[t]
  \centering
  \includegraphics[width=.8\linewidth]{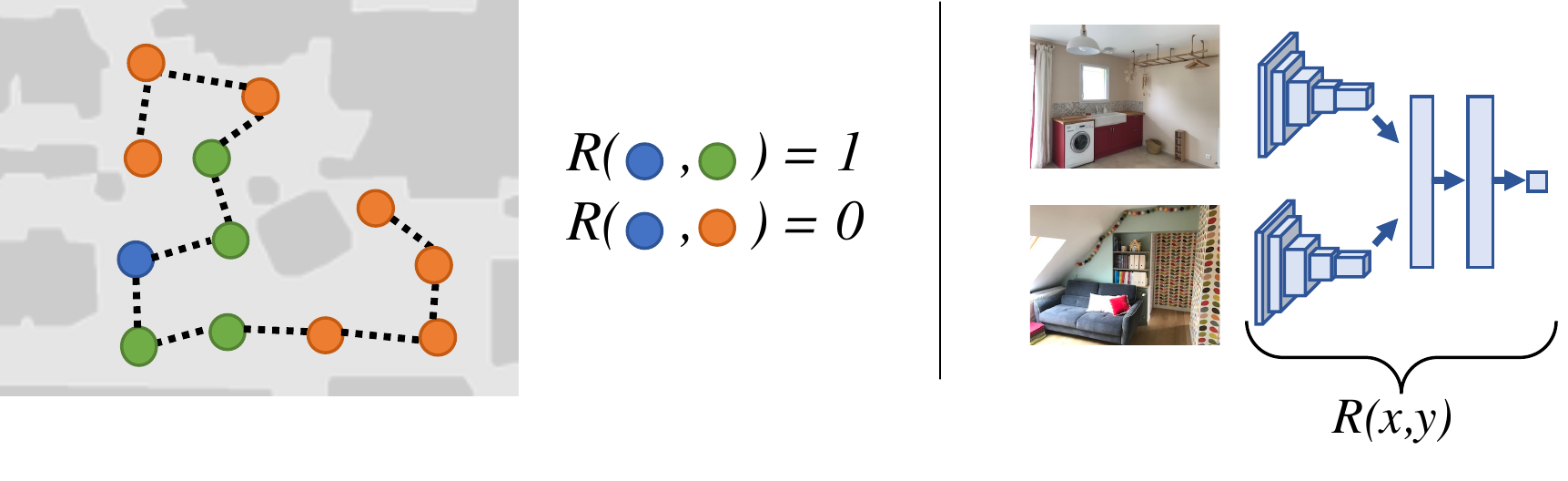}
  \vspace{-3mm}
  \caption{
\textbf{Reachability network~\cite{savinov2018episodic}.}
Given a set of observations made by an agent with a random walk policy \textbf{(left)}, we train the (local) reachability network $R$ to distinguish between observations that are temporally near or distant.
For a given observation (marked in blue), the nearest observations are in green and the distant ones in red.
The reachability network \textbf{(right)} is a siamese network composed of a convolutional network followed by a fully-connected network.
  }
  \label{fig:visualrep}
\end{figure}

As the agent moves around the environment, it receives data from its visual sensor, which in this work produces RGB images.
From this first-person input, the agent builds a representation of its current location that should encode information to distinguish the current location from other locations, as well as give an idea of the distance between locations.
This is  achieved by encouraging nearby locations to have similar representations while pushing distant locations to have different representations.
However, in the absence of information about the agent position or a map, we do not have an explicit notion of distance between locations.

\paragraph{Reachability as an image-based  self-supervision~\cite{savinov2018episodic}.}
An approximation of the spatial distance between locations is the number of time steps taken by an agent with a random walk policy to reach one location starting from the other.
Indeed, the expected distance covered by a random walk is the square root of the number of time steps.
We thus use the temporal distance between observations as a surrogate similarity measure.
More precisely, we let a random agent explore the environment for $T$ steps and collect the sequence of observations, $(\mathbf{x}_1,\dots,\mathbf{x}_T)$.
We then define a reachability label $y_{ij}$ for each pair $(\mathbf{x}_i,\mathbf{x}_j)$ of observations based on their distance in the sequence, i.e., the label $y_{ij}$ is equal to $1$ if $|i-j|\le k$ and $0$ otherwise, $k$ being a hyperparameter.

\paragraph{Learning visual features.}
We train a siamese neural network $R$ to predict the reachability label $y_{ij}$ from the input pair $(\mathbf{x}_i,\mathbf{x}_j)$ with a logistic regression.  It is parameterized by a feedforward network $f$ and a convolutional network $g$ such that $R(\mathbf{x}_i,\mathbf{x}_j)=f(g(\mathbf{x}_i), g(\mathbf{x}_j))$~\cite{savinov2018episodic}.
We use the resulting convolutional network $g$ to form visual features and the siamese network $R$ in the reward function of the exploration module.   This stage is summarized in Fig.~\ref{fig:visualrep}.

\begin{figure}[t!]
  \centering
  \includegraphics[width=.45\linewidth]{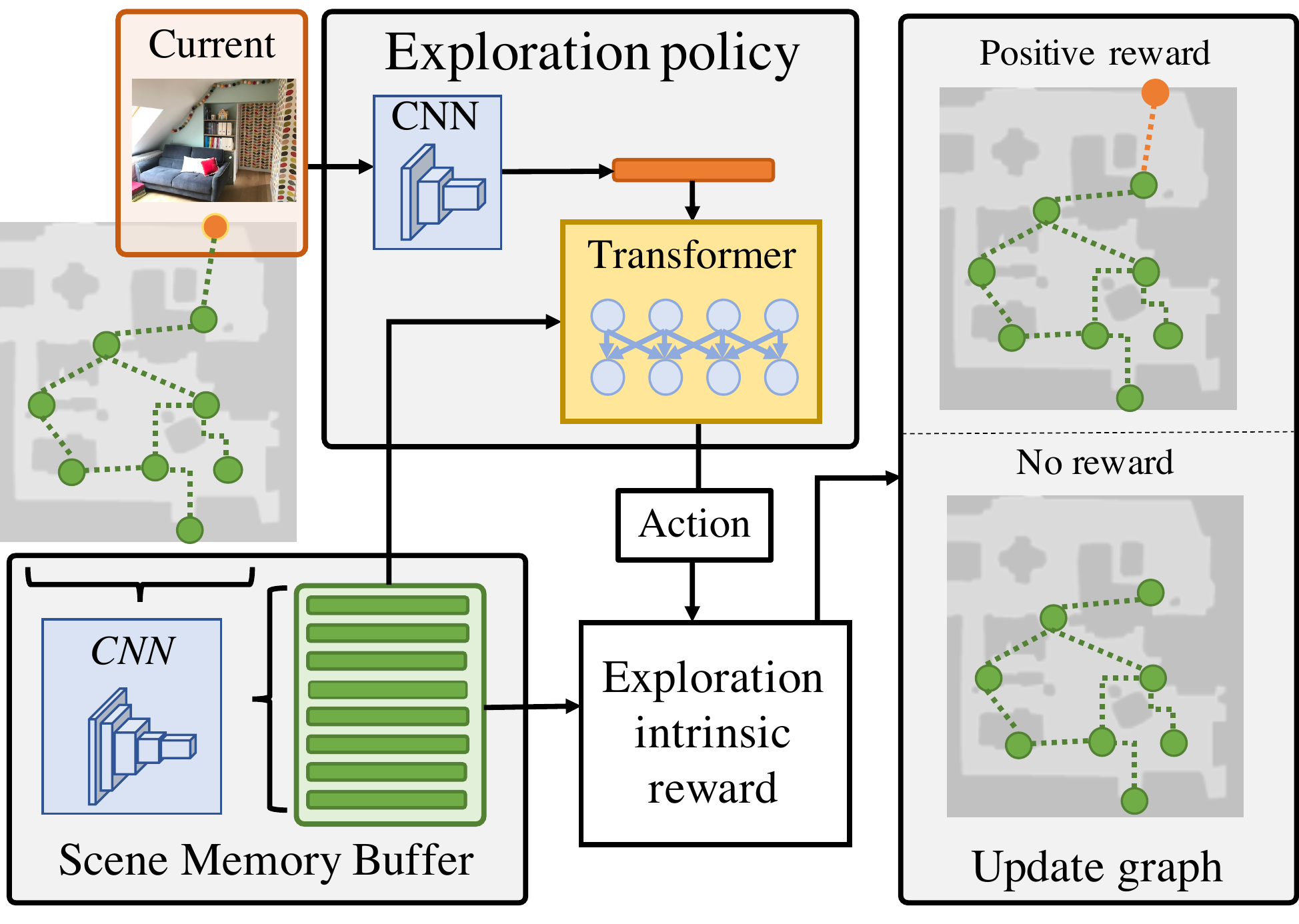} \hspace{5mm}
  \includegraphics[width=.45\linewidth]{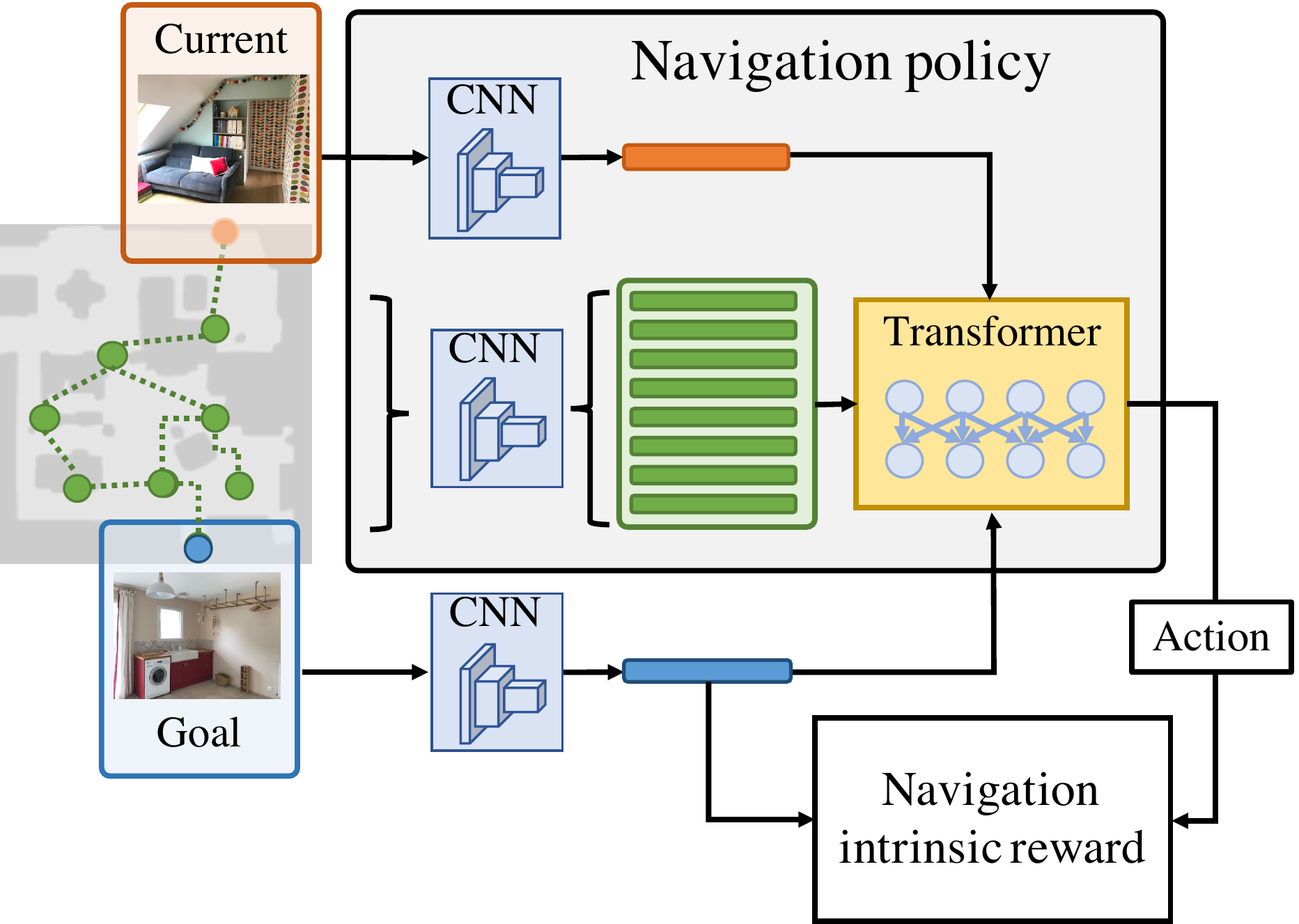}
  \caption{
\textbf{Exploration and navigation stages.} The agent first learns to explore \textbf{(left)} the environment using a Scene Memory Buffer (SMB) of visited regions for its policy and intrinsic reward. 
Next, the agent learns to navigate \textbf{(right)} using SMB to set image oriented goals to itself and learn to navigate towards them. 
  }
  \label{fig:maps}
\end{figure}

\subsection{Stage 2: Learning to Explore}

Once  the agent can differentiate images of nearby locations from distant locations, it can explore and map the environment.
In this section, we describe how to train our exploration module with a curiosity-driven intrinsic reward, which is the second stage of our training.

\subsubsection{Exploration module.}

The agent explores the environment to dynamically build an internal map.
At each step, this map and the current observation are used to plan an action that moves the agent toward unexplored regions.
We model the internal map as a scene memory buffer that contains important past observations, and the agent takes actions by applying a Transformer on this memory buffer. This stage is shown in Fig~\ref{fig:maps}~(left).

\paragraph{Scene Memory Buffer.} 
The agent has a Scene Memory Buffer (SMB) module that stores some of its previous observations.
At each time step $t$, the SMB $\mathbf{M}_{t-1}$ stores an unstructured set of $J_{t-1}$ visual features, $\mathbf{M}_{t-1}=(\mathbf{m}_1,\dots,\mathbf{m}_{J_{t-1}})$.
Storing every observation is not efficient and we follow the mechanism of Savinov~\etal~\cite{savinov2018episodic} to select which observation to store, i.e., $J_{t-1} \le t-1$.
The idea is to add only observations that are distant from the current memory vectors.
Since the siamese network $R$ has been trained explicitly to distinguish close from distant observations, we compute a score of novelty by comparing the current observation with the SMB, i.e.,
\[ s(\mathbf{x}_t,\mathbf{M}_{t-1}) = \max\{ f(g(\mathbf{x}_t),\mathbf{m})~|~\mathbf{m}\in \mathbf{M}_{t-1}\}, \]
and we  propose the following rule to update the SMB:
\begin{equation}
  \mathbf{M}_{t} = \begin{cases}
    \mathbf{M}_{t-1} \cup g(\mathbf{x}_t) & \text{if~~} s(\mathbf{x}_t, \mathbf{M}_t) < \tau, \\
    \mathbf{M}_{t-1} & \text{otherwise},
  \end{cases}
\end{equation}
where $\tau$ is a threshold that influences the radius covered by each memory vector in the SMB.
The SMB will reset after each episode.

\paragraph{Transformer on the SMB.}
The navigation policy  exploits the SMB to move toward unexplored locations through a Transformer.
More precisely, we apply a Transformer layer on top of the SMB and the visual features of the current location to form a vector $\mathbf{h}_t$.
The logits for the navigation policy  and its value function are both linear functions of this vector $\mathbf{h}_t$.
Overall, at time step $t$, the agent receives an observation $\mathbf{x}_t$ and has an SMB $\mathbf{M}_t$.
From those we compute the vector $\mathbf{h}_t$ in the following way:
\begin{align}
\label{eq:exploration}
  \mathbf{c}_t & = \texttt{CNN}(\mathbf{x}_t), \\
  \mathbf{e}_t & = \texttt{LN}(\texttt{Att}(\mathbf{c}_t, \mathbf{M}_{t-1}) + \mathbf{c}_t), \\
  \mathbf{h}_t & = \texttt{LN}(\texttt{MLP}(\mathbf{e}_t) + \mathbf{e}_t),
\end{align}
where \texttt{Att}, \texttt{MLP} and \texttt{LN} denote respectively, the multi-head attention, the feedforward and the layer-normalization sublayers of a Transformer.
Note that the \texttt{CNN} is a convolutional network different from $g$.
We also add absolute temporal position embeddings to encode the temporal distance between the current time step and the moment a memory vector was inserted in the SMB.
We refer the reader to Vaswani~\etal~\cite{vaswani2017attention} for more details on Transformers.

\subsubsection{Instrinsic exploration reward.}
\label{sec:bonus}
Intrinsic curiosity rewards the agent for exploring parts of the environment that looks unfamiliar to the agent.
This reward is based on the agent's intrinsic representation of the environment.
In our case, this representation is the Scene Memory Buffer and a positive reward is given if the current observation has been added to the SMB, i.e.,
\begin{equation}
  r_\text{curiosity}(t) = \alpha \indicator{\{s(\mathbf{x}_t, \mathbf{M}_{t-1}) < \tau\}}.
  \label{eq:bonus}
\end{equation}
This reward is a discrete version of the episodic curiosity~\cite{savinov2018episodic}.
Discretizing the reward removes the trivial solutions noticed in~\cite{savinov2018episodic} where the agent stops in a location that gives a reward that is greater than any reachable locations.

\subsection{Stage 3: Learning to Navigate}
\label{sec:navigation}

In this section, we describe the third stage of our algorithm: the training of our navigation module.
Every episode will start with an exploration phase where the exploration module builds an internal map of the environment. This is followed by a navigation phase that trains the navigation module to reach a goal sampled from the map. The internal map is also used for generating the intrinsic navigation reward.   
The trained navigation module does not need to follow the 
visited locations on the map --- these are only used during training to shape the reward.  In particular, the navigation policy can be more efficient than policies that plan over visited 
locations on the map at test time.  
This stage of the training is depicted in Fig~\ref{fig:maps}~(right).

\subsubsection{Building an internal map.}
In the exploration phase of an episode, an internal map of the environment is built by the exploration policy that is already trained in the previous stage. 
The exploration policy runs for $T$ steps and fills the SMB with visual representations of $J_T$ locations. While the SMB alone is sufficient for training the navigation module with sparse rewards, we also record the connectivity of those $J_T$ locations to be leveraged in the dense-reward version of the training.

The path followed by the agent connects different memory vectors in the SMB.
We use this path to form a graph $G_t$ on top the SMB $\mathbf{M}_t$.
More precisely, we keep track of the closest element $\mathbf{p}_t$ of the current observation $\mathbf{x}_t$ after updating the SMB.
Note that this means that $\mathbf{p}_t$ is equal to $\mathbf{x}_t$ if it is added to the SMB.
If $\mathbf{p}_t$ is different from $\mathbf{p}_{t-1}$, we add an edge $e = (\mathbf{p}_{t-1}, \mathbf{p}_t)$ to $G_t$.
This results in a directed graph representing paths between the memory vectors of the SMB.

\subsubsection{Navigation module.}

The navigation module takes as an input the current observation $\mathbf{x}_t$ as well as a target observation $\mathbf{x}^*$.
The module transforms these observations into features with a CNN, and concatenates the resulting features.
We then apply a Transformer layer on top of this vector and the SMB, resulting in a feature $\mathbf{h}_t$.
We compute the feature $\mathbf{h}_t$ as follows:
\begin{align}
  \mathbf{c}_t & = [\texttt{CNN}(\mathbf{x}_t), \texttt{CNN}(\mathbf{x}^*) ], \\
  \mathbf{e}_t & = \texttt{LN}(\texttt{Att}(\mathbf{c}_t, \mathbf{M}_{t-1}) + \mathbf{c}_t), \\
  \mathbf{h}_t & = \texttt{LN}(\texttt{MLP}(\mathbf{e}_t) + \mathbf{e}_t).
\end{align}
Similar to the exploration module, the policy and value function are linear functions of a feature $\mathbf{h}_t$.
Note that set of parameters for the attention modules for the exploration and navigation modules are different, but not the CNNs.

\subsubsection{Memory based navigation reward.}
\label{sec:graph}

After the exploration phase of an episode, the navigation phase starts by setting a randomly selected element $\mathbf{m}_j$ of the SMB as a goal to navigate towards.
A positive intrinsic reward is given if the agent considers that it has reached the target location based on its reachability network, i.e.,
\begin{equation}
  r_\text{sparse navigation}(t) = \beta \indicator{R(\mathbf{x}_t, \mathbf{x}^*) > \tau}.
\end{equation}
This is an intrinsic reward built solely on the capability of the agent to perceive if it has reached the goal sampled from its SMB.
However this reward is sparse and we propose to densify the reward by further exploiting the SMB.

\paragraph{Dense intrinsic navigation reward.}
We leverage the graph $G_t$ to form dense navigation reward by computing a graph based approximation of the distance to the goal.
More precisely, at each time step $t$, we compute the shortest path between $\mathbf{p}_t$ and $\mathbf{x}^*$ in $G_t$ and denote by $l_t$ its length.
We thus add a dense reward based on this distance as:
\begin{equation}
  r_\text{dense navigation}(t) = \max \left ( 0, \min_{t' < t} \ l_{t'} - l_t \right ).
  \label{eq:dense}
\end{equation}
Note that, since we update the graph $G_t$ as we navigate the environment, this reward may change over time for a same target $\mathbf{x}^*$ and memory vector $\mathbf{p}_t$.
Note that this bonus reward only absolute progress towards the goal and the total reward accumulated over an episode is equal to the length of the shortest path as estimated at the beginning of the episode.
Overall, we use both the dense and sparse reward during the navigation phase.

%% file: experiments.tex
\section{Experimental Evaluation}
In this section we present the empirical evaluation of our model.  We evaluate both the exploration and the navigation modules.
Let us start by describing the data we use and providing technical details of our experimental setup.

\subsection{Datasets.}
For a realistic setup we perform all of our experiments on scenes taken from the Gibson dataset~\cite{xiazamirhe2018gibsonenv}.
We run the simulations for these experiments inside the Habitat-sim framework~\cite{habitat19iccv}.
We have selected a subset of eight scenes from the Gibson dataset, based on the quality of the 3d mesh, surfaces, and the number of floors, following the study presented in~\cite{habitat19iccv}.
The scenes are fairly complex as they have 16 rooms on average spanning multiple floors.
Some statistics for the selected scenes are provided in the supplementary material.
The action set contains three actions: moving forward by one meter, and turning right or left by 45 degrees.
We only keep the RGB data, discarding the depth channel,  and use images of size $160 \times 120$ pixels.

In this work, we make the assumption that the agent is always spawned in the same location of a scene.
To achieve this, for each scene we manually select a starting position corresponding to the entrance door in the house.

\subsection{Implementation Details.}

\paragraph{Visual Representation Learning.}
We implement the network $R$ as a siamese network with resnet18 as the function $g$, and use a comparison function $f$ composed of two hidden layers of dimension $512$.
For each scene, we sample $20$ random trajectories of $20$k steps.
From each trajectory we extract $40k$ pairs, yielding a dataset of $800$k image pairs.
The maximal action distance for a positive pair is set to five steps.
We train this network using SGD for $20$ epochs with a batch size of $128$, a learning rate of $0.1$, a momentum of $0.9$, a weight decay of $10^{-7}$ and no dropout.
We do not share parameters between scenes.

\paragraph{Exploration and Navigation.}
For our CNN module, we use a network with $3$ convolutional layers with kernels of size $[9, 7, 5]$, strides of size $[5, 4, 3]$ and number of channels $[32, 64, 128]$.
For the attention on the memory, we use an attention with two heads, a hidden dimension of $64$ and a feedforward network with a hidden dimension of $128$.
We train the policy using PPO, where each batch consists of 16 full episodes, each with $1000$ steps.
We run $4$ PPO epochs, with $\gamma$ set to $0.99$, an entropy coefficient set to $0.01$ and clipping of $0.1$.
We optimize the parameters using RMSprop with a learning rate of $10^{-4}$, a weight decay of $10^{-7}$, and parameters $\alpha$ and $\epsilon$ set to $0.98$ and $10^{-5}$ respectively.
For this model we do use dropout with $p=0.1$, and a learning rate warm up phase of $300$ steps.
As with the $R$ network, we do not share parameters between scenes.


\subsection{Main Results}
\label{sec:navi_result}
The main experiment in our evaluation checks how well our agent navigates to new test goals.
After training itself to navigate to elements of the memory, the agent can be given a new goal feature as a target.
In this experiment, we sampled 100 random locations from each scene, and saved the corresponding RGB observation and location.
For each scene, and each target location, we first run 1000 steps of exploration to fill the memory and launch the navigation episode.
The total navigation episode lasts for 1000 steps, and as soon as a goal is reached, a new goal is sampled.

\paragraph{Evaluation Metrics.}
We evaluate success by computing the number of targets that the agent reached successfully out of 100.
The target is considered reached if the agent navigates to a distance of at most one meter from the target location.
The first metric that we compute is the \emph{success rate}, which simply corresponds to the fraction of goals that were reached within the allocated 1000 steps.
Since this measure does not account for the length of the path taken, the second metric we report is the SPL metric~\cite{anderson2018evaluation}.
Let us assume that we have access to the length of the shortest path from the starting location to the goal $i$, computed by the simulator, that we denote $l_i$.
If we write $s_i$ the indicator of success as defined above, and $d_i$ the metric length of the trajectory obtained with our algorithm, the SPL is defined as follows:
\begin{equation}
  SPL = \frac{1}{N} \sum_{i=1}^N s_i \ \frac{l_i}{\max(l_i, d_i)}.
\end{equation}

\paragraph{Baselines.}
In order to evaluate the quality of our navigation algorithm, we compare our model to three baselines: SPTM, Supervised and Random.
We describe these baselines in more detail here.

First, we compare our algorithm to the Semi-Parametric Topological Memory~\cite{savinov2018semi}.
In order to adapt SPTM to the environments used in our experiments, we train the action and edge prediction networks on them.
For each scene, we train the networks for 300 epochs of 1000 batches each, with a batch size of 64.
Samples in the batches are obtained from random trajectories that are sampled on-line.
This number of training iterations amounts to approximately $90$M steps in each environment - which is comparable to the number of steps used to train our method (exploration and navigation).
Since SPTM requires an expert human-provided exploration trajectory, we use random exploration in place.

\begin{table}[t]
  \centering
  \caption{
    Navigation performance as measured by the SPL metric for our method (Ours) and selected baselines on all considered environments.
  }
  \setlength\tabcolsep{1mm}
  \begin{tabular}{@{}l c c c c c c c c c c@{}}
    \toprule
                    & Adrian     & Albert.      & Arkan.      & Ballou    & Capist.     & Goffs  & Mosq.    & Sanc.        && Mean    \\
                    \midrule
    Random                      & 13.5 & 19.3 & 16.4 & 10.5 & 26.0 &  9.3 & 10.6 & 12.6 && 14.8 \\ 
    SPTM~\cite{savinov2018semi} & 25.5 & 23.5 & 20.2 &  9.7 & 38.6 &  9.3 & 16.9 & 10.1 && 19.2 \\
    Supervised                  & 27.5 & 30.5 & 21.9 & 11.1 & 45.8 & 14.8 & 13.0 & 17.4 && 22.8 \\
    \midrule
    Ours-sparse                 & 27.8 & 39.9 & 30.6 & 17.0 & 60.9 & 15.1 & 16.3 & \textbf{32.9} && 30.1 \\
    Ours-dense          & \textbf{35.6} & \textbf{45.2} & \textbf{32.3} & \textbf{27.8} & \textbf{65.9} & \textbf{16.8} & \textbf{18.8} & 24.5 && \textbf{33.4} \\
    \bottomrule
    \end{tabular}
  \label{tab:navigation-spl}
\end{table}

Second, we also compare against a feedforward policy trained with supervised rewards (\emph{Supervised}).
This policy is trained using RL, assuming that at each step the distance $d(t)$ from the agent to the goal is known: $d(t) = \|p_\text{agent}(t) - p_\text{goal}(t)\|_2$.
In that setup, the agent receives a reward of 10 if $d(t) < 1$ - which is equivalent to the success criterion defined above.
Please note that this feedforward policy is trained on the same set of 100 goals that are used during evaluation.
For reference, we provide the performance of random navigation.

\begin{figure}[t]
  \centering
  \includegraphics[width=0.31\linewidth]{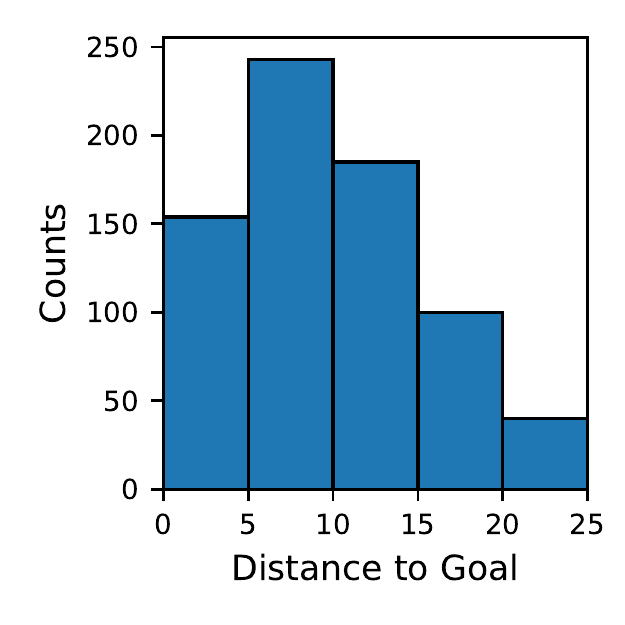}
  \includegraphics[width=0.31\linewidth]{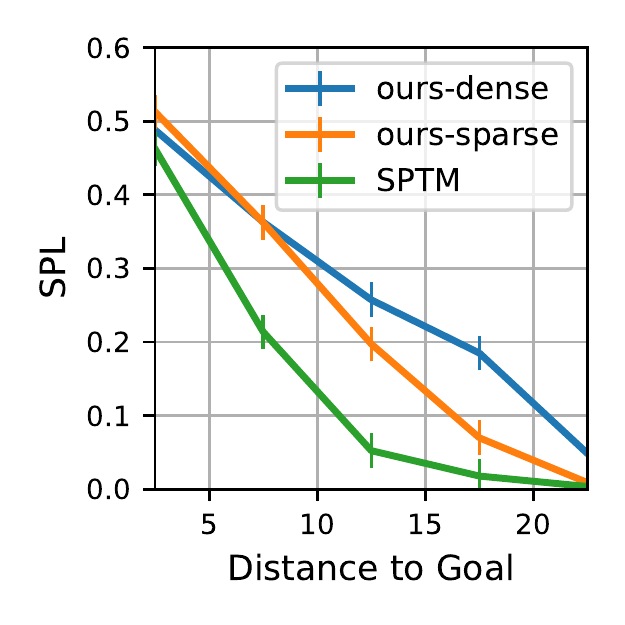}
  \includegraphics[width=0.31\linewidth]{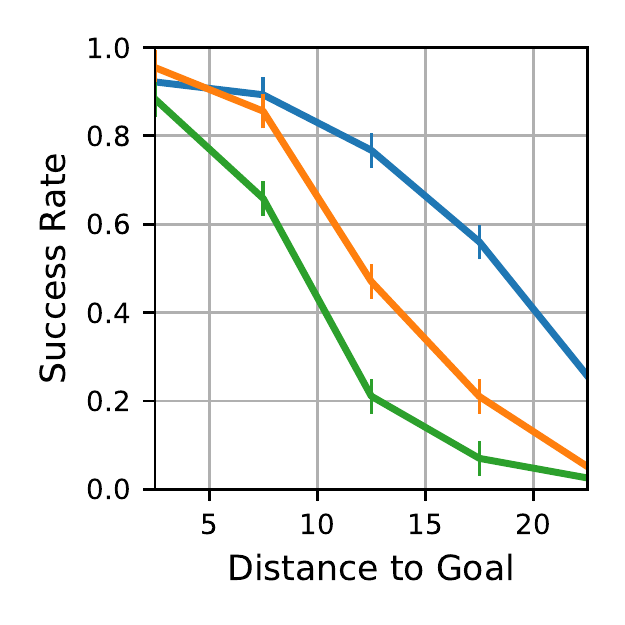}
  \label{fig:nav-measure}
  \caption{
    Navigation performance is broken down by physical shortest distance from start to goal location.
    \textbf{Left}: Histogram of distances from start to goal in our evaluation dataset. We see that most goals are located between $5$m and $10$m from the entrance.
    \textbf{Center}: Breakdown of the SPL metrics by distance. 
    \textbf{Right}: Breakdown of the flat success rate by distance.
    We clearly see the advantage of using the dense rewards for learning to navigate to far-away locations.
  }
\end{figure}

\paragraph{Results.}
We run the evaluation for the baselines and our method with sparse or dense rewards and report the results for each scene in Table~\ref{tab:navigation-spl}.
There are a couple observations that we can make about this experiment.
First of all, we see that our method outperforms all the baselines by a large margin on all of the scenes.
Surprisingly, it even works better than the supervised agent which utilizes the location information - data to which our method has no access.
However, this can be explained by our architectural choice of conditioning the navigation module on the memory of previously visited states.
As opposed to that, the \emph{Supervised} baseline is only a feedforward network, and has no representation of the past observations.

Second, the SPTM baseline performs poorly compared to our method with only little improvement over the randomly acting agent.
This can be explained by the fact that SPTM only has access to a random exploration trajectory, therefore limiting the set of goals that it can ever reach.
Moreover, SPTM restricts the navigation to its exploration graph, severely limiting the possible routes to the goal.
In comparison, our method encourages the agent to reach the goal as fast as possible taking any possible route. 
Our dense reward does use the graph, but only as a guide that can be completely ignored if more optimal solutions exist.

Finally, we see that the dense reward generated using the graph $G_t$ as described in Sec.~\ref{sec:graph} allows to train a better navigation policy, outperforming the sparse reward on most of the scenes.
Indeed, for our agent, this dense reward corresponds to a discrete distance over the graph which leads to the goal if minimized. 
This effect is clearer when we measure the performance for different goal distances as shown in Fig.~\ref{fig:nav-measure}.
The gap between the dense and sparse reward widens for far-away goals.
This is likely because the graph $G_t$ provides intermediate goals, helping a lot when the goal cannot be reached easily.


\subsection{Analysis of Exploration}

As mentioned in Sec.~\ref{sec:navigation}, the coverage obtained during the exploration stage is critical for the final navigation task.
In this section, we want to evaluate the quality of this exploration stage alone.

\paragraph{Evaluation Metrics.}
The goal of the exploration stage is to train an agent to explore and map an environment without any form of supervision.
For this experiment, we follow previous work and evaluate the quality of the exploration using a coverage metric.
In order to define this metric, we discretize the environment using a grid, with cells of size $1 \times 1$ meter.
At the end of the episode, we report the number of cells that were visited by the agent.
Since the environments we consider can have multiple floors, we infer the floors in the environment.
We do so by sampling random locations and keeping most frequent heights that are at least $.5$ meters away by doing non-maximal suppression.
We then keep one coverage grid per floor.

\paragraph{Baselines.}
First, we compare our exploration module to Episodic Curiosity \emph{(EC)}~\cite{savinov2018episodic}.
In that baseline - unlike our method - the policy has no dependency on the past.
This is implemented by making the policy and value function directly depend on $c_t$ instead of $h_t$ in Eq.~(\ref{eq:exploration}).
Another difference between our method and the \emph{EC} baseline, is the nature of the intrinsic rewards.
While the original bonus proposed in~\cite{savinov2018episodic} was continuous, we propose to use a discretized version instead.
Note that we cannot compare to EC on the navigation task in Sec.~\ref{sec:navi_result} because it does not provide means of navigation without supervision.

Second, we include a \emph{Supervised} policy trained using the ``oracle'' reward, being the measure that we use for evaluation.
In this case, we densely reward the agent every time a new cell is visited.
Apart from using a different source of reward, all parameters for this model are taken the same as for our model.

\begin{figure}[t]
  \centering
  \includegraphics[width=0.33\linewidth,trim=0 10 0 0,clip]{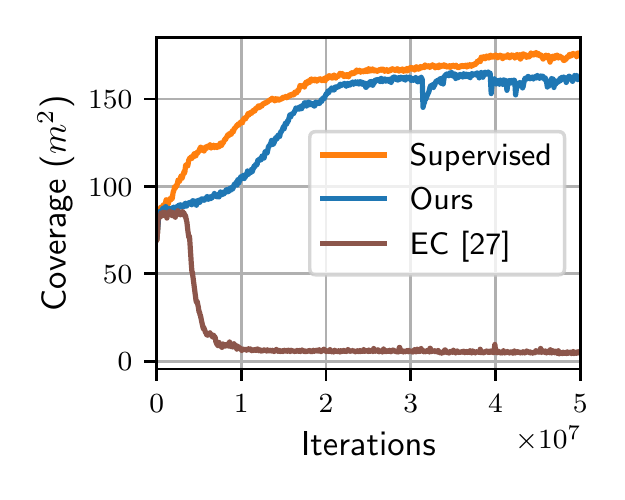}
  \includegraphics[width=0.65\linewidth,trim=0 0 0 0,clip]{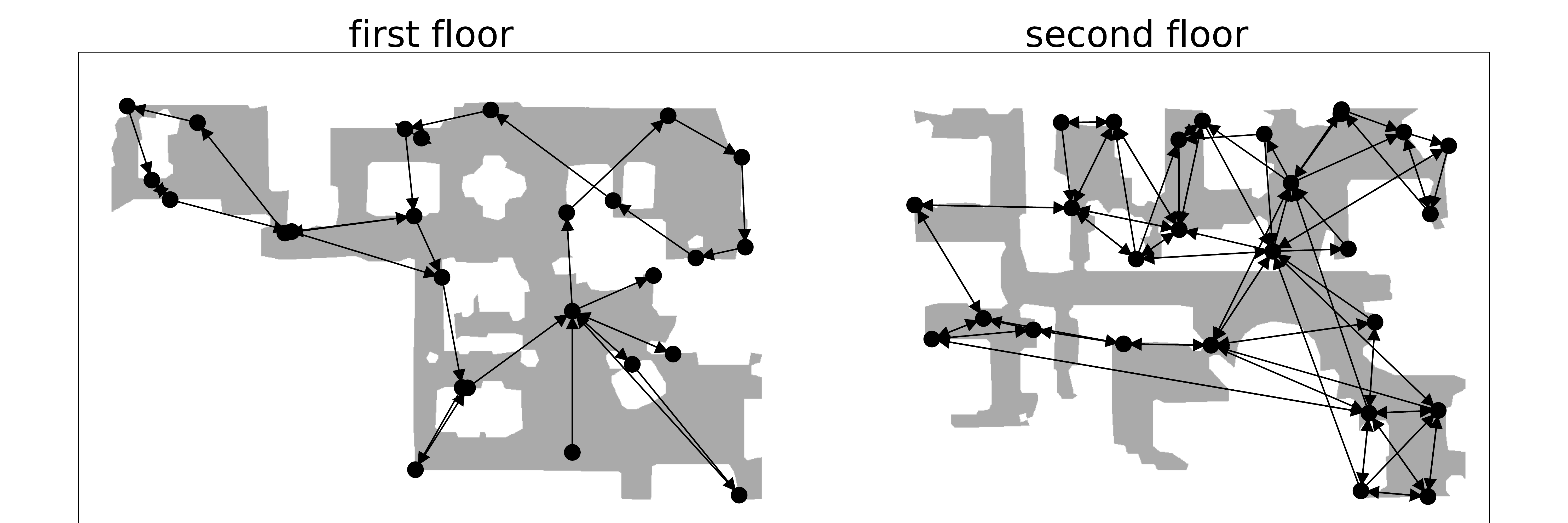}
  \caption{
    Quantitative and qualitative evaluation of the exploration phase.
    \textbf{Left}: Performance of exploration policies as measured by the coverage metric in squared meters.
    We compare the performance of our model to a baseline (EC) and a supervised topline.
    \textbf{Right}: Visualization of the graph build during exploration in the Ballou environment.
  }
  \label{fig:oracle-measure}
\end{figure}

\paragraph{Results.}
The performance evolution of our method and the baselines during training is shown in Fig.~\ref{fig:oracle-measure}~(Left). The coverage metric is averaged over the eight scenes.
Our method performs comparable to the supervised agent, which can be considered as the upper bound as it directly optimizes the coverage metric.
In Fig.~\ref{fig:oracle-measure}~(Right), we show an example of exploration behavior learnt by our agent. 
The nodes of this graph are states added to the SMB by the agent, and they are connected following the rule described in Sec.~\ref{sec:navigation}.
We see that the agent has explored most of the house successfully and made connections consistent with its topology, which will assist the training of the navigation module.  
Surprisingly, we observed that the agent trained using vanilla EC does not learn a good exploration policy.
We investigate the reason for this in the following experiment.

\begin{figure}[t]
  \centering  
  \begin{tabular}{c c c c}    
  & & \scriptsize Continuous & \scriptsize Discrete \\
  \includegraphics[height=9.5em,trim=0 5 0 0,clip]{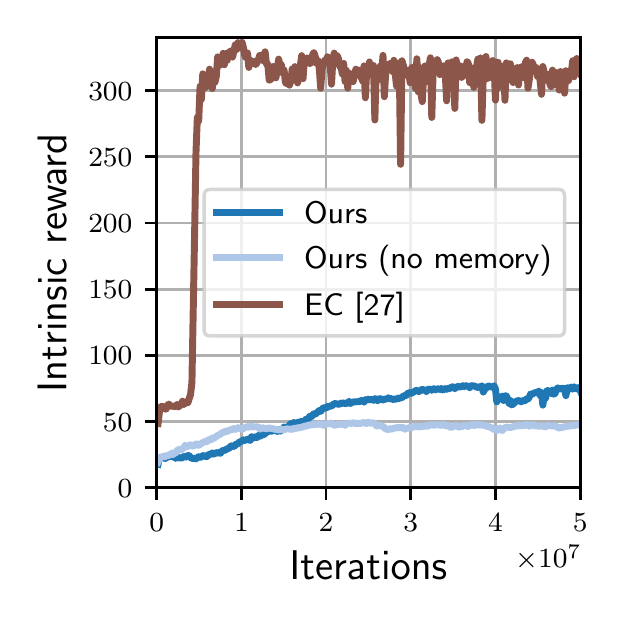} & 
  \includegraphics[height=9.5em,trim=0 5 0 0,clip]{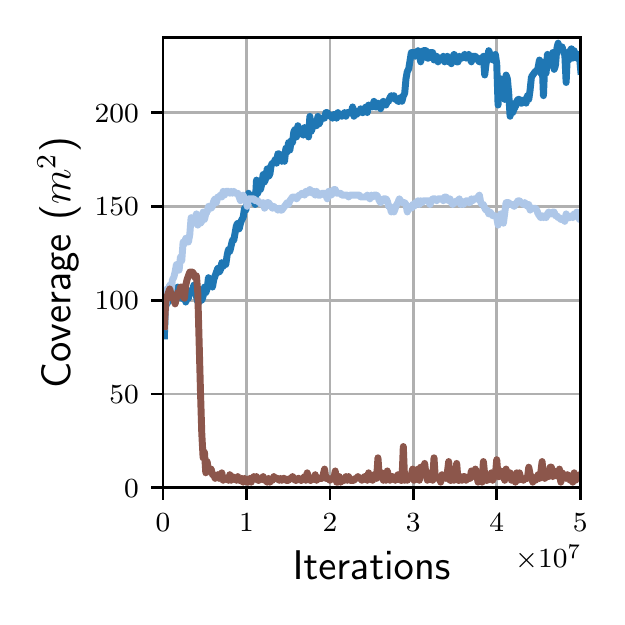} &
  \includegraphics[height=9.5em,trim=0 0 0 85,clip]{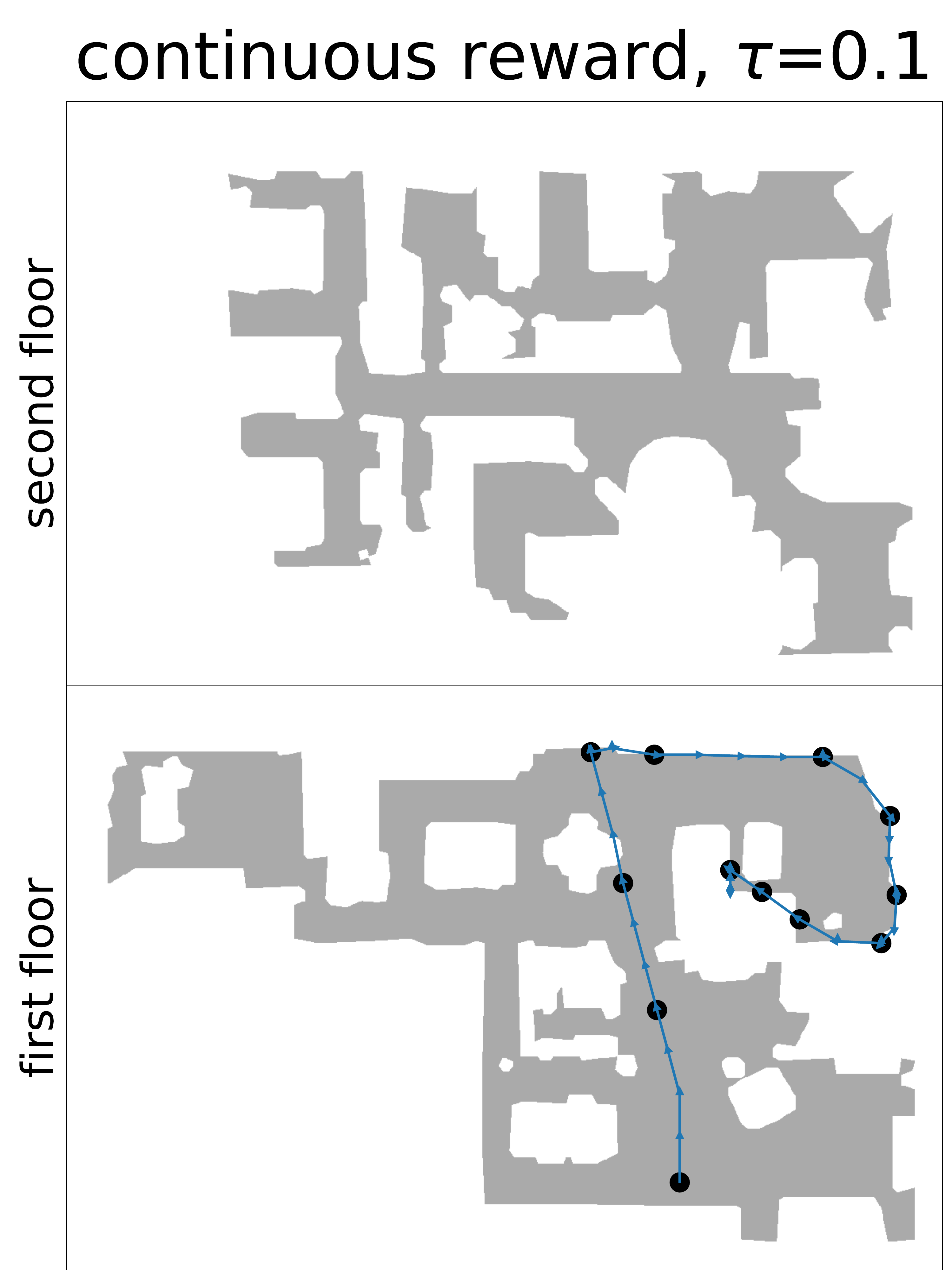} &
  \includegraphics[height=9.5em,trim=0 0 0 85,clip]{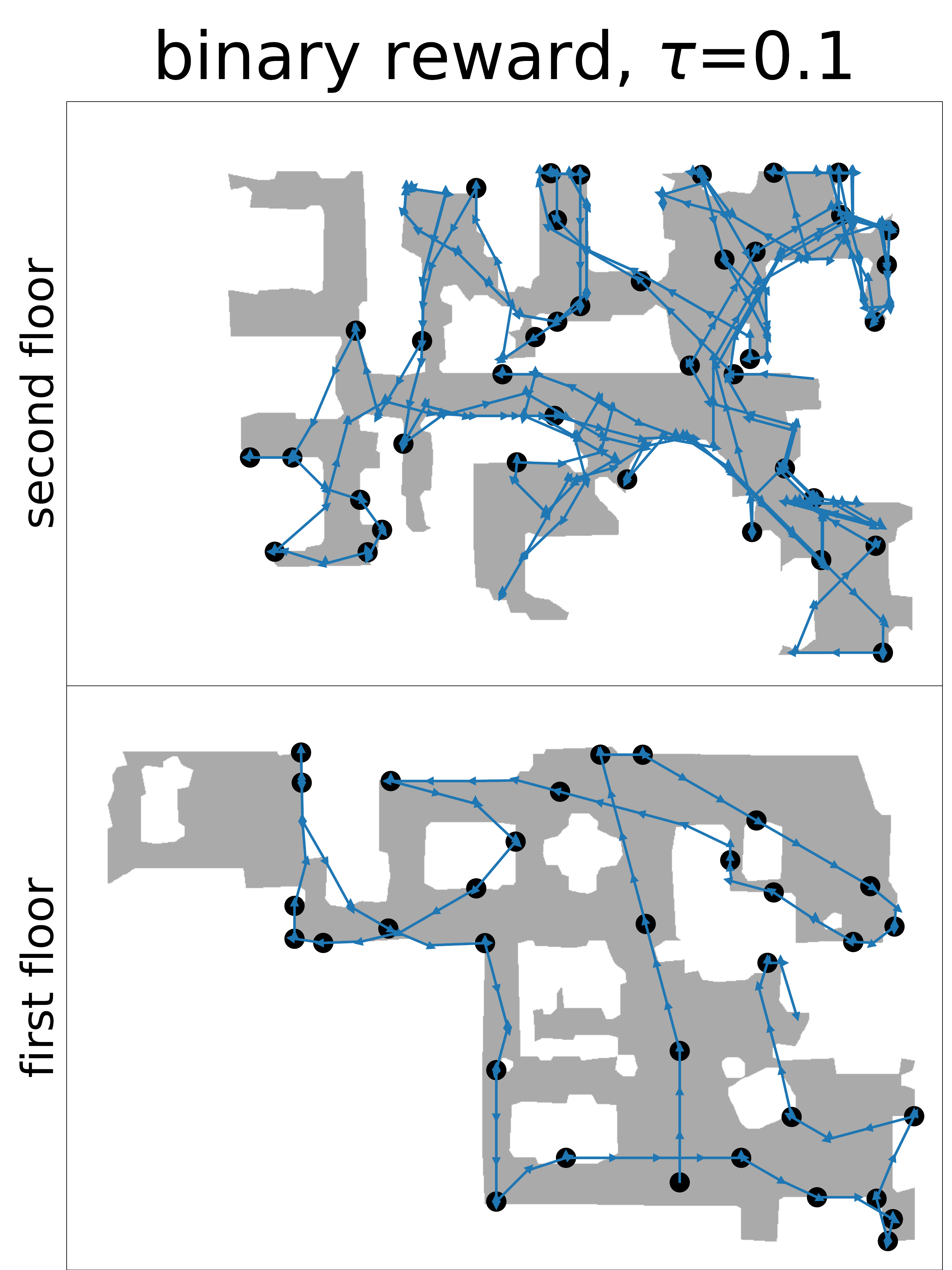} 
  \end{tabular}
  \caption{
    Ablation study of our exploration policy.
    We report the performance for our model, the EC baseline, as well as a variant of our model with no attention mechanism on the SMB.
    \textbf{Left}: Evolution of the intrinsic bonus reward as a function of iterations.
    \textbf{Center}: Evolution of the coverage metric.  
    \textbf{Right}: Visualization of the trajectories obtained with a policy trained with continuous and discrete bonus rewards.
  }
  \label{fig:binary-reward}
\end{figure}

\paragraph{Ablation of the exploration model.}
In~\cite{savinov2018episodic}, the authors propose a continuous curiosity reward: $r(t) = \alpha . (\beta - s(\mathbf{x}_t, \mathbf{M}_{t-1} ))$.
In this ablation study, we want to exhibit the importance of our improvements over~\cite{savinov2018episodic}, namely using a discrete bonus and an attention mechanism over the SMB.
To this end, we show the evolution of the intrinsic reward as well as of the coverage metric for three models on the Ballou environment.
We compare our full model to the vanilla EC and an exploration policy such as ours but with no memory in Fig.~\ref{fig:binary-reward}.

We observe that using the continuous reward makes the agent find trivial maximas by exploiting the reward design.
In that case the total episode reward converges to a value just below $N \times \tau$, where $N$ is the number of steps - see Fig.~\ref{fig:binary-reward}~(Left).
Despite the fact that the agent trains properly, and optimizes the reward, it does not work well when measured by the metric we care about, the coverage metric, as shown in Fig.~\ref{fig:binary-reward}~(Center).
We provide a qualitative representation of the phenomenon, by visualizing the agent's path, as well as the spatial location of elements in the memory.
We show these visualizations for both continuous and discrete rewards in Fig.~\ref{fig:binary-reward} (Right).
We see that the agent trained with discrete rewards manages to explore the scene properly.
However, when trained with continuous intrinsic rewards, the agent gets stuck in a specific subpart of the environment where it receives a continuous reward just below the threshold $\tau$.

